\documentclass{article}

\usepackage[final]{neurips_2024}
\usepackage{hyperref}       
\usepackage{graphicx}
\usepackage{amsmath}
\usepackage{multirow}

\usepackage[utf8]{inputenc} 
\usepackage[T1]{fontenc}    

\usepackage{url}            
\usepackage{booktabs}       
\usepackage{amsfonts}       
\usepackage{nicefrac}       
\usepackage{microtype}      
\usepackage{xcolor}         

\title{Language-Guided Trajectory Traversal in Disentangled Stable Diffusion Latent Space for Factorized Medical Image Generation}

\author{%
  Zahra ~TehraniNasab$^*$\\
  McGill University\\
  MILA-Quebec AI Institute\\
  \texttt{zahra.tehraninasab@mail.mcgill.ca} \\
  \And
  Amar ~Kumar$^*$\\
  McGill University\\
  MILA-Quebec AI Institute\\
  \texttt{amar.kumar@mail.mcgill.ca} \\
  \And
  Tal ~Arbel\\
  McGill University\\
  MILA-Quebec AI Institute\\
  \texttt{tal.arbel@mcgill.ca} \\
}

\begin{document}

\maketitle
\renewcommand{\thefootnote}{\fnsymbol{footnote}}
\footnotetext[1]{Equal contribution.}
\begin{abstract}
Text-to-image diffusion models have demonstrated a remarkable ability to generate photorealistic images from natural language prompts. These high-resolution, language-guided synthesized images are essential for the explainability of disease or exploring causal relationships. However, their potential for disentangling and controlling latent factors of variation in specialized domains like medical imaging remains under-explored. In this work, we present the first investigation of the power of pre-trained vision-language foundation models, once fine-tuned on medical image datasets, to perform latent disentanglement for factorized medical image generation and interpolation.   Through extensive experiments on chest X-ray and skin datasets, we illustrate that fine-tuned, language-guided Stable Diffusion inherently learns to factorize key attributes for image generation, such as the patient's anatomical structures or disease diagnostic features. We devise a framework to identify, isolate, and manipulate key attributes through latent space trajectory traversal of generative models, facilitating precise control over medical image synthesis.
\end{abstract}

\section{Introduction}
\label{sec:intro}

Deep learning models for medical imaging have shown state-of-the-art performance across several tasks such as disease classification, image segmentation, drug discovery and high-resolution image synthesis. However, these models often struggle to generalize well to new, unseen data due to domain shifts~\cite{quinonero2022dataset} or entanglement in image features~\cite{meng2020learning}. 
For example, in medical imaging, multiple factors, such as disease pathology and imaging modality, are often intertwined, making it difficult for the model to isolate the relevant features for the specified task. This entanglement can lead to the model relying on spurious correlations, which may not be present in new data, thus hindering its ability to generalize. Thus, disentanglement aims to separate the various factors of variation in the data, allowing the model to learn more robust and interpretable representations.  In computer vision, disentanglement models~\cite{higgins2017beta,kim2018disentangling,chen2018isolating} have been shown to help increase generalization~\cite{long2015learning,tzeng2017adversarial} and explainability~\cite{soelistyo2024discovering} by isolating task-relevant features from confounding factors. This enables the model to adapt to new data distributions. In addition, disentanglement improves model explainability by separating the factors of variation and providing a clearer understanding of how different features contribute to the final prediction.

\begin{figure}
    \centering
    \includegraphics[width=0.95\linewidth]{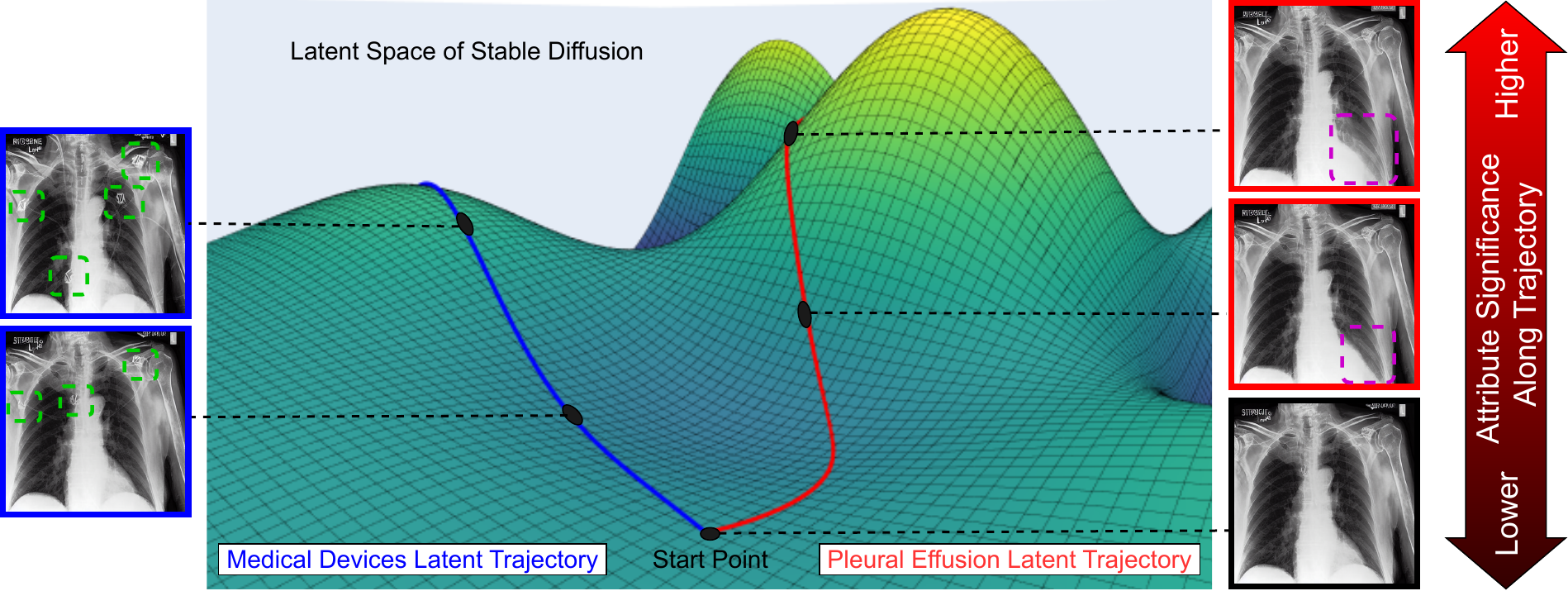}
    \caption{Traversal along the latent trajectories of Stable Diffusion using language guidance. Given an initial chest X-ray projected onto latent space (start point), traversal along the trajectory is performed via language guidance. Sampling along the trajectory results in only a single attribute (e.g. ``medical devices", ``pleural effusion") being altered from the start point (``neutral"), while the patient identity is maintained.}
    \label{fig:intro}
\end{figure}
Vision-language foundation models have emerged as a powerful approach for learning disentangled representations, offering several advantages over traditional generative architectures due, in large part, to the enormous dataset size they have been trained on. Earlier approaches designed to disentangle latent factors relied on specialized generative architectures such as Variational Auto Encoders (VAE)~\cite{leeb2020structured}, which don't produce high-resolution images required for medical imaging, or Generative Adversarial Networks (GAN)~\cite{chen2016infogan}, which are notoriously hard to train. Normalizing Flows~\cite{pawlowski2020deep} have been adapted to address disentanglement in simpler contexts but are prohibitively computationally expensive. 
Other challenges faced when adapting traditional models include their difficulties in training them end-to-end~\cite{tang2021disentangled,zhou2022lung}, and their requirement for specialized architectural components for conditioning such as AdaIN~\cite{huang2017arbitrary}, FiLM~\cite{perez2018film}, SPADE~\cite{park2019semantic}. Furthermore, models often require specific heuristic to permit traversals in latent space~\cite{harkonen2020ganspace,shen2020interpreting,shen2020interfacegan,shen2021closed}. In contrast, vision-language foundation models offer more efficient and scalable solutions. 
The power of these foundation models to disentangle latent representations, enabling targeted image modifications while preserving semantic content and other attributes, has been explored in the natural imaging domain~\cite{wu2023uncovering,wu2023not}. However, this domain remains under-explored in the context of medical images, where complex entanglement is common and challenging to address.

In this paper, we investigate the disentanglement capabilities of Stable Diffusion~\cite{Rombach_2022_CVPR}, fine-tuned on medical images, and propose the first method to traverse latent space based on language guidance and see the factorizing effect on the resulting images.
Language guidance is seen to be effective at identifying \textit{attribute-specific trajectories} in the latent space (see Figure~\ref{fig:intro}). Additionally, interpolations between samples permit continuous trajectories that can be sampled. Experiments illustrate that the samples exhibit the same disentangled properties, that is, the attribute of interest remains the same in the image throughout the trajectory, becoming more prevalent proportional to the distance from the starting point.  We propose a new metric, {\it Classifier Flip Rate} along a Trajectory (CFRT), to validate the presence of the desired disentanglement.  

\section{Methodology}
\label{sec:methodology}
\begin{figure}[t]
    \centering
    \includegraphics[width=0.95\linewidth]{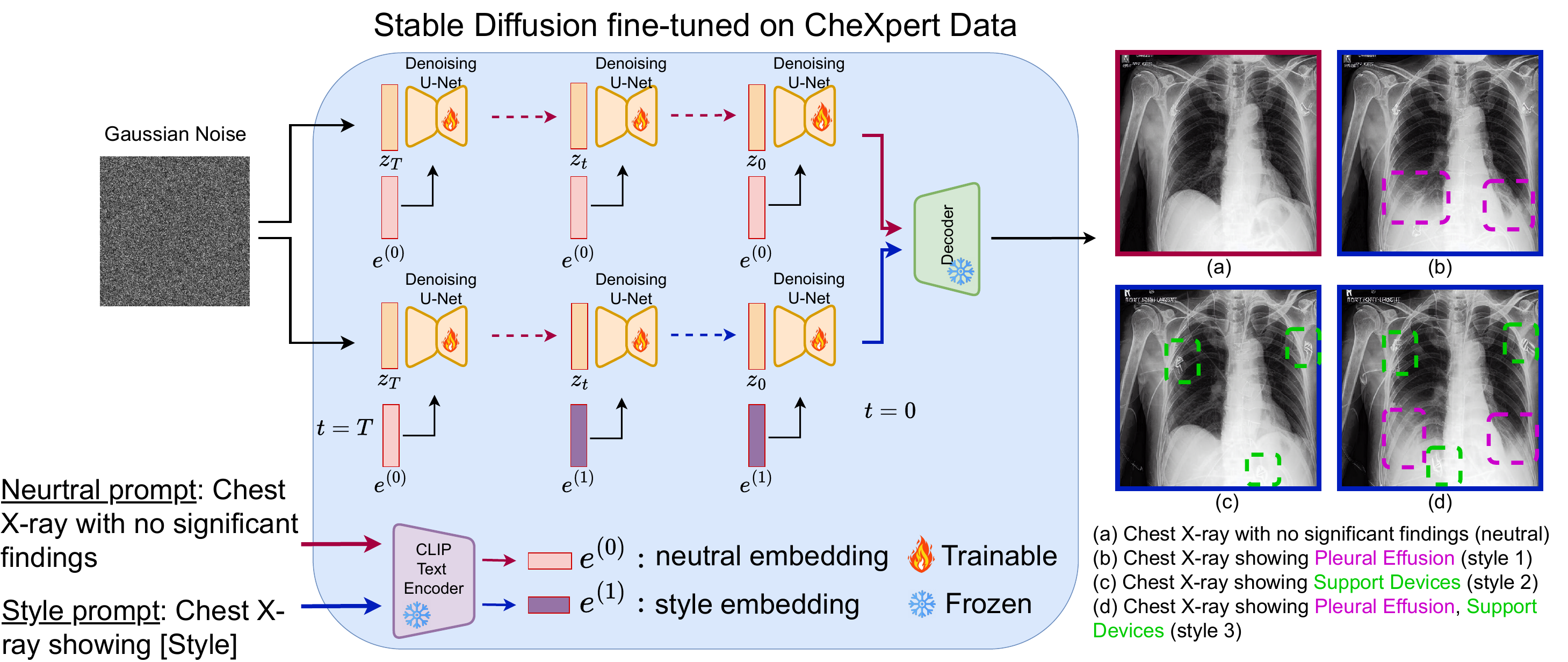}
    \caption{Reverse Diffusion for synthesizing disentangled images from text prompts using fine-tuned Stable Diffusion. The reverse diffusion process takes noisy latents, $z_T$, as input. A U-Net architecture generates the denoised latent, $z_0$. During the denoising process, the text embeddings, $e_{t}$ ($t\in{0,..,T}$), from the pre-trained CLIP encoder are added to the latent via cross-attention modules~\cite{Rombach_2022_CVPR}. Finally, the de-noised latent is passed through the decoder to create the synthesised image. Note that the text embeddings can be replaced at some intermediate timestep $t$ during the reverse diffusion.}
    \label{fig:architecture}
\end{figure}

Our method for building disentangled representations follows a two-stage process: (i) fine-tune the pre-trained Stable Diffusion on the relevant dataset (in this paper, CheXpert or ISIC); (ii) denoising diffusion implicit model (DDIM)~\cite{song2020denoising} conditioned on the text embedding to synthesize images along the trajectory (see Figure~\ref{fig:architecture}).
\subsection{Text-to-Image Generation using fine-tuned Stable Diffusion}
Given a text embedding, $e$, the objective of the text-conditioned fine-tuned DDIM is to generate an image, $X$. The forward diffusion process starts at the timestep $t=0$ and noises the image $X_0$ till timestep $t=T$ resulting in a set of noisy images, $X_{1:T}$. The denoising process of the DDIM attempts to denoise from $X_T$ to the original image $X_0$ where a noisy image at each step is given by $X_{t-1}$:  
\begin{equation}\label{eq:denoising}
    X_{t-1} = \gamma_{t_0}X_t + \gamma_{t_1}\epsilon_0\left(X_t,t,e_t\right).
\end{equation}
Here, $\gamma_{t_0}$ and $\gamma_{t_1}$ are computed as $\gamma_{t_0} = \sqrt{\frac{\alpha_{t-1}}{\alpha_t}}$ and $\gamma_{t_1} = \sqrt{1-\alpha_{t-1}} - \sqrt{\frac{\alpha_{t-1}}{\alpha_t} - \alpha_{t-1}}$ respectively with $\alpha_{1:T}$ as the hyperparameters for the diffusion process.
In the usual setting of text-to-image generation, a single text prompt can generate a single image. In this work, image generation depends on the diffusion timestep, $t$, to explore the properties of the latent representation. Thus, the image generation process, $X_0$,~\cite{wu2023uncovering} can be described as in Eq~\ref{eq:geneartion}, where $\mathbf{g}$ is the generator:
\begin{equation}\label{eq:geneartion}
    X_0 = \mathbf{g}(X_T, e_{1:T}).
\end{equation}

\textbf{Fine-tuning Stable Diffusion: }Similar to~\cite{kumar2025PRISM}, CheXpert's binary labels are converted into text captions to fine-tune Stable Diffusion v1.5~\cite{Rombach_2022_CVPR}, which utilizes CLIP language embeddings rather than numerical labels for as conditioning. The models are trained using text prompts with the following template: \texttt{Chest X-ray of a subject with [finding(s)]} or \texttt{a dermoscopic image with [disease] showing [artifacts]} for CheXpert or ISIC respectively (see Section~\ref{dataset_implementation}). To ensure stability while fine-tuning, the CLIP encoder and VAE encoder/decoder remain frozen and train the denoising U-NET component. This approach preserves the CLIP encoder's semantic knowledge while adapting the image generation to match the CheXpert data distribution. 
\\\noindent\textbf{Inference:} A ``neutral" image, $X_0$ is first generated based on a given text prompt, with embeddings $e$. To add specific attributes, such as disease(s) or style(s) such as hairs and gel bubbles, to the ``neutral" image, a new text prompt with embeddings $e'$ is used. During the reverse diffusion, $e'$ replaces the original text embeddings $e$ at some time-step $t$. Replacing the embedding at different timesteps results in variations in the denoised latent, $z_T$. The properties of these images are further discussed in section~\ref{sec:results}.

For a given attribute associated with chest X-rays, such as the appearance of support devices (e.g. pacemaker, wires, tubes)  or disease states (e.g. pleural effusion), the disentangled latent representations of the attributes of high-dimensional medical images should follow a non-linear trajectory. With a discrete set of images, a smooth interpolation between samples is made possible through B\'ezier curves~\cite{mortenson1999mathematics}, which would be preferable over the more choppy linear interpolation~\cite{chen2018b}.
\subsection{Metrics for evaluating the conditionally generated images \& attribute disentanglement}


We hypothesize that the disentanglement of attributes exists along different trajectories in the latent space of Stable Diffusion, where the latent codes fall along a nonlinear trajectory. We propose a new metric,~\textit{Classifier Flip Rate along a Trajectory (CFRT)}, to validate disentanglement along the specified (style) trajectory. In Equation~\ref{eq:cfrt}, the CFRT score ensures that an attribute classifier's decision changes more significantly from the starting point (``neutral image") when the attribute of interest, $\mathcal{A}$, is modified compared to changes from modifying any other attribute.
\begin{equation}\label{eq:cfrt}
\begin{split}
\text{CFRT}_{\mathcal{A}} =\\\frac{1}{|X|} \sum_{x \in X} \mathbb{1} \bigg[ 
&|f(x) - f(x_{\mathcal{A}}')| > \max_{j \neq \mathcal{A}} |f(x) - f(x_j')|\\
&\wedge \ y(x) = y(x_{\mathcal{A}}') \wedge \ \forall k \neq \mathcal{A}, x_k = x_{\mathcal{A}}'(k) \bigg]
\end{split}
\end{equation}
where $X$ is the set of all the samples, $x_\mathcal{A}'$ is the conditionally generated sample, $x$ is the starting point in the trajectory, where only attribute $\mathcal{A}$ is flipped, $f(.)$ is the classifier's prediction for sample $x$, $y(x)$ is the ground truth label for sample $x$, $\mathbb{1}[\cdot]$ is the indicator function with value 1 if the condition is true and 0 otherwise and $\forall j \neq \mathcal{A}, x_j = x_\mathcal{A}'(j)$ ensures that all attributes $j$ other than $\mathcal{A}$ remain the same between $x$ and its counterfactual $x_\mathcal{A}'$.
A higher value of CFRT indicates attribute-specific changes to the image without affecting the other attributes, thus showing disentanglement. 

The trajectory can be identified using a text prompt indicating an attribute of interest (e.g. ``pleural effusion"). A second metric, the pairwise \textit{cosine similarity}~\cite{liu2019latent} (between the directional vectors), can be used to prove that the latent representations for a particular attribute lie on a non-linear trajectory (thus cannot be captured easily by Style-GAN's~\cite{lang2021explaining} singular value decomposition which assumes that the trajectory is linear). More formally, the cosine similarity between the direction of the latent representation of the conditionally generated ($z_{cg}$) image at timestep $t_i$ ($z_{cg,t_i}$) and at timestep $t_j$ ($z_{cg,t_j}$), both relative to the latent of the original (``neutral") image, $z_\text{orig}$.
\begin{equation}\label{eq:cosine}
    \cos(\theta_{t_i, t_j}) = \frac{\Delta\vec{z}_{t_i} \cdot \Delta\vec{z}_{t_j}}{||\Delta\vec{z}_{t_i}|| \cdot ||\Delta\vec{z}_{t_j}||}, \quad \forall t_i,t_j \in T
\end{equation}
where $\Delta\vec{z}_{t} = \vec{z}_{\text{cg},t} - \vec{z}_{\text{orig}}$ and $T = \{5k : k \in \{1,2,...,9\}\}$.
Finally, the visual quality of the conditionally generated images can be validated using a standard metric, ~\textit{Learned Perceptual Image Patch Similarity (LPIPS)}~\cite{zhang2018unreasonable}. This metric can be computed for the conditionally generated samples, as well as any samples obtained through interpolation along an attribute trajectory, relative to the starting point. A lower LPIPS score indicates that two images are more similar/closer in perceptual quality.

\section{Experiments \& Results}
\subsection{Dataset and Implementation Details}\label{dataset_implementation}
We perform experiments on two publicly available datasets - CheXpert~\cite{irvin2019chexpert} and ISIC2019~\cite{tschandl2018ham10000, codella2018skin, combalia2019, bcn20000}. 
For CheXpert, \texttt{findings} include pleural effusion or support devices while for ISIC2019, \texttt{disease} includes melanoma (MEL) and melanocytic nevus (NV). We use the artifact attributes, such as hair, gel bubbles, ink and ruler, provided by Bissoto et al.~\cite{bissoto2022artifact}\footnote{Artifact information for ISIC is available at \url{https://github.com/alceubissoto/artifact-generalization-skin}. }. The number of samples for train, validation and test splits are shown in Appendix~\ref{appen:num} 

\texttt{Stable-Diffusion-v1-5}~\cite{Rombach_2022_CVPR} is used for text-to-image generation (high-resolution, $512\times512$) tasks. The pre-trained model is frozen, except for the U-net in the denoising process (see Figure~\ref{fig:architecture}) throughout all experiments. A variant of the DDIM sampler~\cite{liu2022pseudo} is used for generating images with 50 total backward diffusion steps (for latent samples along the trajectory, we chose the timesteps from $t=5$ to $t=45$ in steps of 5). We use 7 sample points to approximate a B\'ezier curves~\cite{mortenson1999mathematics} and sample 50 new latents along this interpolated trajectory to evaluate if these samples exhibit the same properties as their trajectory. For better stability, the variance of the additive Gaussian noise is set to 0~\cite{kim2022diffusionclip}. 
\subsection{Results}\label{sec:results}
\begin{figure}[t]
    \centering
    \includegraphics[width=0.95\linewidth]{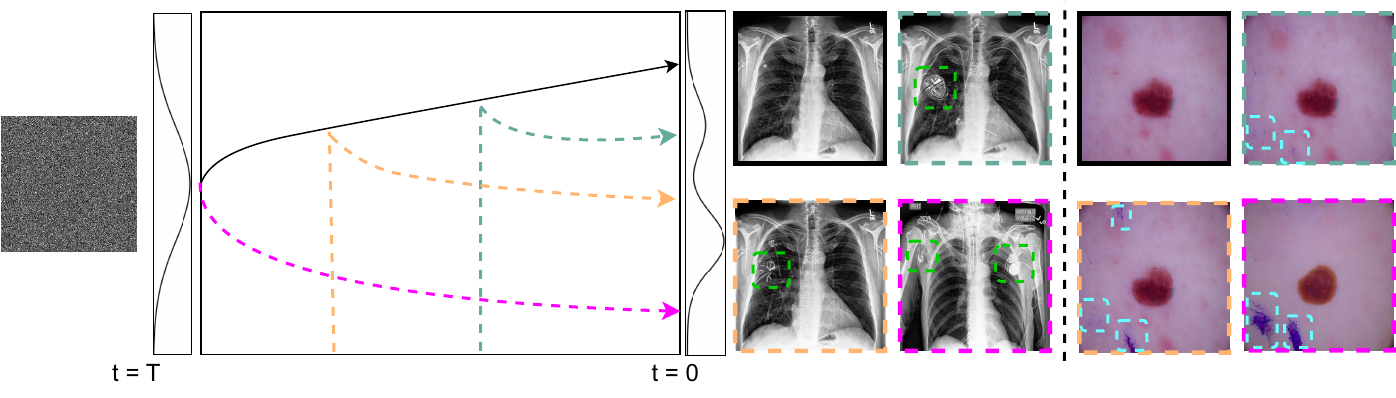}
    \caption{Disentanglement property of the Stable Diffusion~\cite{wu2023uncovering}. Starting from Gaussian noise (left image) at sampling timepoint t=T, the reverse diffusion process denoises the image (right) at timepoint t=0. The text prompts for the ``neutral" images (with \textbf{dark borders}) for CheXpert and ISIC are \texttt{Chest x-ray with no significant findings} and \texttt{A dermoscopic image with melanocytic nevus (NV)}, respectively. The images on the right (matched with coloured borders) are the synthesized images with the same text prompts \texttt{Chest x-ray showing Support Devices} for CheXpert and \texttt{a dermoscopic image with melanocytic nevus (NV) showing ink} sampled at different timesteps during the reverse diffusion process. Notice that sampling closer to the timepoint t=0 results in a synthesized image similar to the \textbf{original image} and as we sample closer to the timepoints t=T, the patient's anatomical structure changes. }
    \label{fig:qual_results1}
\end{figure}

\noindent\textbf{\underline{Qualitative evaluations}:} The trajectory in Stable Diffusion latent space corresponding to the desired attribute can be empirically traversed using text representing the target attribute, such as diseases or artifacts. First, we establish the disentanglement capability of text-to-image generation in vision-language foundation models between style (attribute) and content (general anatomy of the patient in the image). Empirically, swapping the initial text prompt (neutral) with the new text prompt (with style) at different time steps in the reverse diffusion process results in variations of the same attribute without changing the content (patient) or other attributes (disease/artifacts) - see Figure~\ref{fig:qual_results1}. 

\begin{figure}[t]
    \centering
    \includegraphics[width=0.95\linewidth]{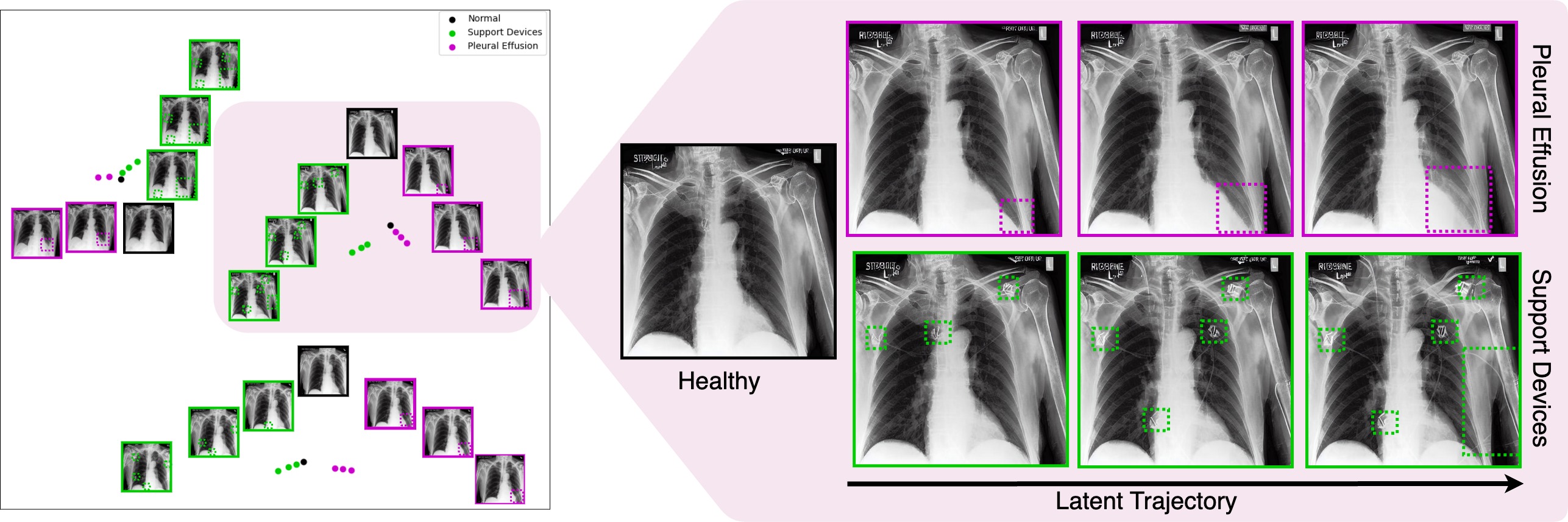}
    \caption{t-SNE plot of generated latent vectors of Stable Diffusion sampled from noise showing disentanglement. The dots and the images with borders show the resulting Stable Diffusion latent vectors and their corresponding "neutral" images with the text prompt - \texttt{Normal chest x-ray with no significant findings}. For each generated sample, we swap the original text condition to \texttt{Chest x-ray showing Support Devices} for one trajectory and to \texttt{Chest x-ray showing Pleural Effusion} for a different trajectory at multiple denoising steps during reverse diffusion.}
    \label{fig:qual_results2}

\end{figure}

Next, we explore the properties of the latent vectors of the conditionally generated images from Stable Diffusion for different attributes. Empirically, for each specific subject, there exist trajectories in the latent space that correspond to specific attributes. The direction along these trajectories can be discovered using an input text prompt. As we traverse in these directions, the desired attribute becomes more prominent in the image without affecting the other confounding attributes, indicating the disentanglement of stable diffusion latent space. This can be seen in  Figure~\ref{fig:qual_results2}. To validate our hypothesis further, we also perform a curve fitting through these latent points and sample new points on this interpolated curve. The properties of the images from these interpolated points affirm disentangled trajectories for multiple attributes (shown in Appendix~\ref{appen:interpolations}). 

\begin{figure}[h]
    \centering
    \includegraphics[width=0.95\textwidth]{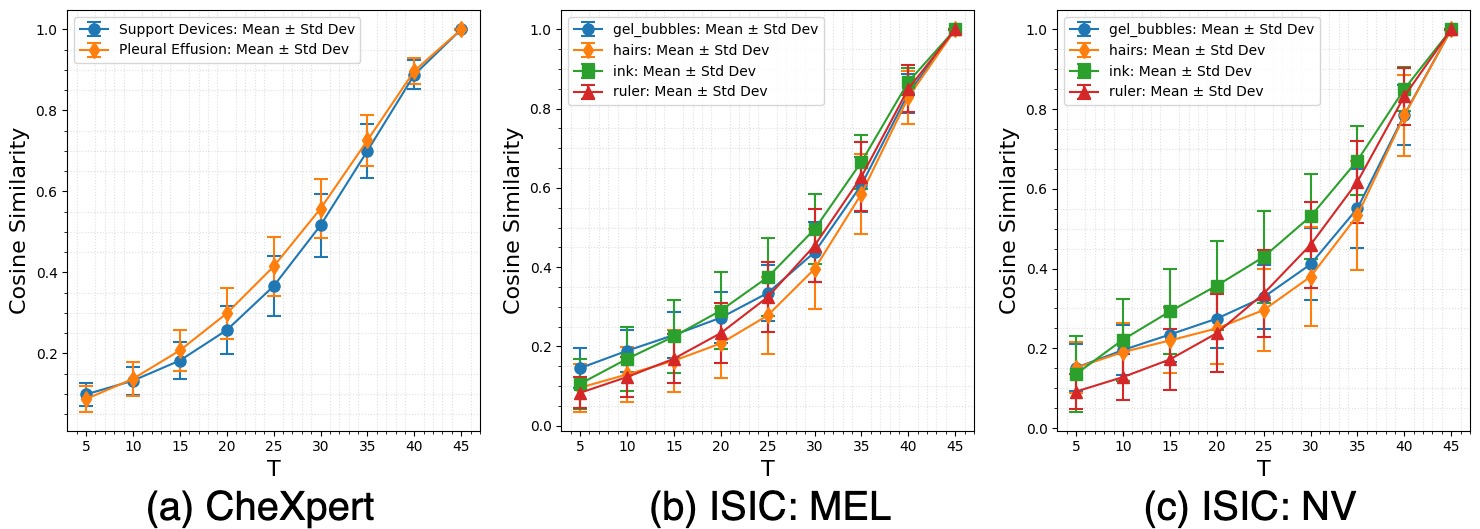}
    \caption{Cosine similarity for all the attributes.}
    \label{fig:cosine}
\end{figure}

\noindent\textbf{\underline{Quantitative evaluations}:} 
 In Table~\ref{tab:quantitative}, the CFRT and the LPIPS scores between the ``neutral image" and the conditionally generated style images at different timesteps along the trajectory (including interpolated samples) are high and low, respectively. This indicates that the attribute of interest along a trajectory changes more than other attributes, and all the interpolated samples show visual similarity to the neutral image. Table~\ref{tab:classifiers} shows the performance of the classifier Efficient-Net~\cite{tan2019efficientnet} for measuring CFRT score. Additionally, the cosine similarities, in Figure~\ref{fig:cosine}, indicate the non-linearity of trajectories for different attributes.   
\begin{table}
\centering
\caption{Performance of the classifiers for identifying disease or artifacts. }

\begin{tabular}{ccc|ccccc}
\hline
 & \multicolumn{2}{c|}{\textbf{CheXpert}} & \multicolumn{5}{c}{\textbf{ISIC}} \\ \cline{2-8} 
\textbf{} & \begin{tabular}[c]{@{}c@{}}Support \\ Devices\end{tabular} & \begin{tabular}[c]{@{}c@{}}Pleural\\ Effusion\end{tabular} & \begin{tabular}[c]{@{}c@{}}MEL /\\ NV\end{tabular} & Hair & \begin{tabular}[c]{@{}c@{}}Gel\\ Bubbles\end{tabular} & Ink & Ruler \\ \hline
\multicolumn{1}{c|}{Accuracy} & 0.86 & 0.80 & 0.91 & 0.93 & 0.94 & 0.96 & 0.97 \\
\multicolumn{1}{c|}{F1-score} & 0.88 & 0.79 & 0.88 & 0.91 & 0.78 & 0.89 & 0.88 \\ \hline
\end{tabular}

\label{tab:classifiers}
\end{table}
\begin{table}
\caption{Quantitative results to evaluate the synthesized images (2500 samples per sub-class). Note: Neutral images for CheXpert are always healthy subjects without support devices. MEL and NV indicates the ``neutral image" with MEL or NV.}
\centering

\begin{tabular}{cccccccc}
\hline
 & \multicolumn{2}{c|}{\textbf{CheXpert}} & \multicolumn{5}{c}{\textbf{ISIC}} \\ \hline
\multicolumn{1}{c|}{Style$\rightarrow$} & \begin{tabular}[c]{@{}c@{}}Pleural\\ Effusion\end{tabular} & \multicolumn{1}{c|}{\begin{tabular}[c]{@{}c@{}}Support \\ Devices\end{tabular}} & & Hair & \begin{tabular}[c]{@{}c@{}}Gel \\ Bubbles\end{tabular} & Ink & Ruler \\ \hline
\multicolumn{1}{c|}{\multirow{2}{*}{CFRT$\uparrow$}} & \multirow{2}{*}{0.78} & \multicolumn{1}{c|}{\multirow{2}{*}{0.89}} & MEL & 0.91 & 0.99 & 0.59 & 0.74 \\
\multicolumn{1}{c|}{} &  & \multicolumn{1}{c|}{} & NV & 0.86 & 0.97 & 0.71 & 0.95 \\ \hline
\multicolumn{1}{c|}{\multirow{2}{*}{LPIPS$\downarrow$}} & \multirow{2}{*}{0.24} & \multicolumn{1}{c|}{\multirow{2}{*}{0.05}} & MEL & 0.08 & 0.09 & 0.12 & 0.11 \\
\multicolumn{1}{c|}{} &  & \multicolumn{1}{c|}{} & NV & 0.05 & 0.09 & 0.06 & 0.10 \\ \hline
\multicolumn{8}{c}{Interpolations} \\ \hline
\multicolumn{1}{c|}{\multirow{2}{*}{CFRT$\uparrow$}} & \multirow{2}{*}{0.73} & \multicolumn{1}{c|}{\multirow{2}{*}{0.86}} & MEL & 0.88 & 0.99 & 0.62 & 0.79 \\
\multicolumn{1}{c|}{} &  & \multicolumn{1}{c|}{} & NV & 0.93 & 0.99 & 0.72 & 0.97 \\ \hline
\multicolumn{1}{c|}{\multirow{2}{*}{LPIPS$\downarrow$}} & \multirow{2}{*}{0.22} & \multicolumn{1}{c|}{\multirow{2}{*}{0.04}} & MEL & 0.05 & 0.08 & 0.09 & 0.08 \\
\multicolumn{1}{c|}{} &  & \multicolumn{1}{c|}{} & NV & 0.04 & 0.07 & 0.04 & 0.07 \\ \hline
\end{tabular}

\label{tab:quantitative}
\end{table}

\section{Conclusion}
Vision-language foundation models have rich latent representations that can be leveraged in medical imaging, where the data are limited. In this paper, we presented the first text-guided traversal along a non-linear trajectory in the disentangled latent space based on vision-language foundation models for medical images. The qualitative and quantitative results demonstrate that our method enables precise control over the synthesized images (including the interpolated images), allowing for targeted manipulation of visual attributes while preserving content. 
Future work will investigate ways to impose more structure and compositionality to the latent spaces.

\newpage
\begin{ack}
The authors are grateful for funding provided by the Natural Sciences and Engineering Research Council of Canada, the Canadian Institute for Advanced
Research (CIFAR) Artificial Intelligence Chairs program, Mila - Quebec AI Institute, Google Research, Calcul Quebec, and the Digital Research Alliance of
Canada.
\end{ack}

{
    \bibliographystyle{ieeenat_fullname}
    \bibliography{main}
}

\clearpage
\appendix

\setcounter{section}{0}

\section{Train, Validation and Test split}
\label{appen:num}
The distribution of samples for both datasets across different splits is shown in Table~\ref{table:samples}.%

\begin{table}[!htb]
\caption{Number of samples in the train, validation and test splits for both datasets. Note that the artifacts: ink has a lower prevalence compared to others.}
\centering
\begin{tabular}{c|ccc}
\hline
& Train & Validation & Test \\ \hline
\multicolumn{4}{c}{CheXpert} \\ \hline
\begin{tabular}[c]{@{}c@{}}Pleural Effusion\end{tabular} & 62509 & 10996 & 12972 \\
\begin{tabular}[c]{@{}c@{}}Support Devices\end{tabular} & 78211 & 13678 & 16196 \\ \hline
\multicolumn{4}{c}{ISIC 2019} \\ \hline
Melanoma & 2750 & 454 & 537 \\
\begin{tabular}[c]{@{}c@{}}Melanocytic Nevus\end{tabular} & 9254 & 1665 & 1956 \\
Hair & 4514 & 802 & 989 \\
\begin{tabular}[c]{@{}c@{}}Gel Bubbles\end{tabular} & 1300 & 228 & 257 \\
Ink & 201 & 34 & 37 \\
Ruler & 1608 & 308 & 341 \\ \hline
\end{tabular}
\label{table:samples}
\end{table}

\begin{figure}[!htbp]
        \begin{minipage}{\textwidth}
            \centering
            \includegraphics[width=\linewidth]{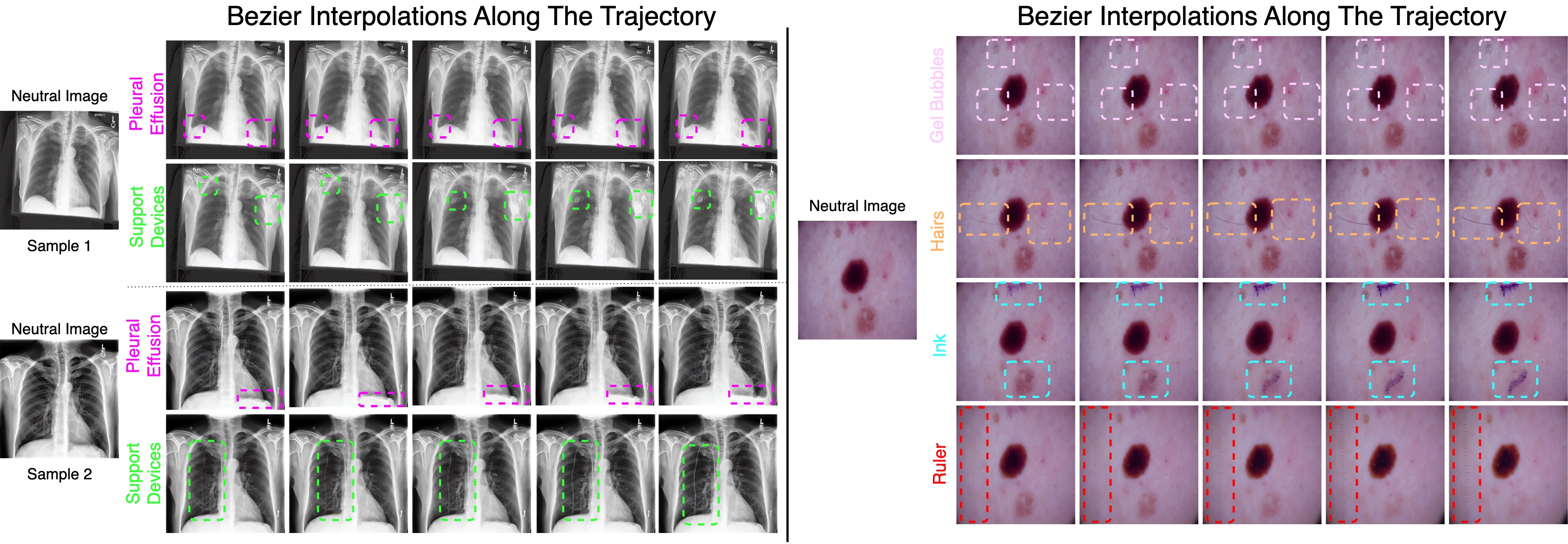}
            \caption{Traversal along the latent trajectories of Stable Diffusion using language guidance. Given a neutral image of a chest X-ray projected onto latent space (start point), traversal along the trajectory is performed via language guidance. Sampling along the trajectory results in only a single attribute (e.g. \textcolor[HTML]{33FF33}{support devices}, \textcolor[HTML]{FF00FF}{pleural effusion}, \textcolor[HTML]{FF99CC}{gel bubbles}, \textcolor[HTML]{FFB366}{hairs}, \textcolor[HTML]{33FFFF}{ink}, \textcolor[HTML]{FF0000}{ruler} ) being altered from the start point (``neutral image"), while becoming more severe along each trajectory, while the patient identity is maintained.}
            \label{fig:interpolations}
        \end{minipage}
\end{figure}

\section{Bézier interpolations}\label{appen:interpolations}
Our method employs Bézier curve interpolations for smooth attribute trajectory modeling~\cite{shimagaki2023bezier} for CheXpert (left image) and ISIC (right image) data. Given a set of points (i.e, conditionally generated style latents: $P_0, P_1, \dots, P_n$), the Bézier curve of degree $n$ is defined as a linear combination of these points weighted by Bernstein polynomials~\cite{lorentz2012bernstein}, defined as:
\begin{equation*}\label{equation}
    B(t) = \sum_{i=0}^{n} \binom{n}{i} (1 - t)^{n-i} t^i P_i, \quad t \in [0,1]
\end{equation*}

\section{Interpolated samples}\label{appen:intepolations}
 B\'ezier curve interpolations are performed on the sample points generated by swapping the text embeddings of the ``neutral image'' with those of the ``style-conditioned text prompt'' during the reverse diffusion process. Figure~\ref{fig:interpolations} shows the interpolated images along the style trajectories for both datasets.





\end{document}